\documentclass[10pt,twocolumn,letterpaper]{article}

\usepackage{cvpr}              

\usepackage{times}
\usepackage{epsfig}
\usepackage{graphicx}
\usepackage{amsmath}
\usepackage{amssymb}
\usepackage{booktabs}
\usepackage{subcaption}

\usepackage[pagebackref,breaklinks,colorlinks]{hyperref}

\usepackage[capitalize]{cleveref}
\crefname{section}{Sec.}{Secs.}
\Crefname{section}{Section}{Sections}
\Crefname{table}{Table}{Tables}
\crefname{table}{Tab.}{Tabs.}

\def\cvprPaperID{13} 
\def\confName{CVPM}
\def\confYear{2023}

\begin{document}

\title{Promoting Generalization in Cross-Dataset Remote Photoplethysmography}

\author{Nathan Vance, Jeremy Speth, Benjamin Sporrer,
Patrick Flynn\\
The University of Notre Dame\\
Notre Dame, IN 46556 USA\\
{\tt\small \{nvance1, jspeth, bsporrer,
flynn\}@nd.edu}
}

\maketitle

\begin{abstract}

Remote Photoplethysmography (rPPG), or the remote monitoring of a subject's heart rate using a camera, has seen a shift from handcrafted techniques to deep learning models. While current solutions offer substantial performance gains, we show that these models tend to learn a bias to pulse wave features inherent to the training dataset. We develop augmentations to mitigate this learned bias by expanding both the range and variability of heart rates that the model sees while training, resulting in improved model convergence when training and cross-dataset generalization at test time. Through a 3-way cross dataset analysis we demonstrate a reduction in mean absolute error from over 13 beats per minute to below 3 beats per minute. We compare our method with other recent rPPG systems, finding similar performance under a variety of evaluation parameters.

\end{abstract}
\section{Introduction}

Measuring a subject's heart rate is an important component of physiological monitoring. While methods such as photoplethysmography (PPG) exist for contact heart rate monitoring, a push has been made for non-contact remote photoplethysmography (rPPG). rPPG is cheaper, requiring a commodity camera rather than a specialized pulse oximeter, and it is contact-free, allowing for applications in new contexts.

Initial techniques for rPPG employed hand crafted algorithms involving a multi-stage pipeline~\cite{DeHaan2013,Wang2017}. While these techniques can be highly accurate, their performance is adversely affected by dynamics common in videos such as motion and illumination changes. More recently, deep learning methods have been applied to rPPG, many of them outperforming the hand crafted techniques~\cite{chen2018deepphys,Yu2019,tsou2020siamese,song2021pulsegan,lu2021dual,liu2020multi}.

While deep learning techniques have benefits, they suffer drawbacks as well in terms of generalization. It has been shown that the learned priors in deep learning rPPG models are strong enough to predict a periodic signal in situations where a periodic signal is not present in the input~\cite{lin2019face} --- a relevant attack scenario. We demonstrate that a deep learning rPPG model may be biased toward predicting heart rate features such as the frequency bands and rates of change that appear in its training data, and therefore struggle to generalize to new situations. We argue that more emphasis on cross-dataset generalization, \ie domain shift, is needed in rPPG research.

\begin{figure}
    \centering
    \includegraphics[width=\linewidth]{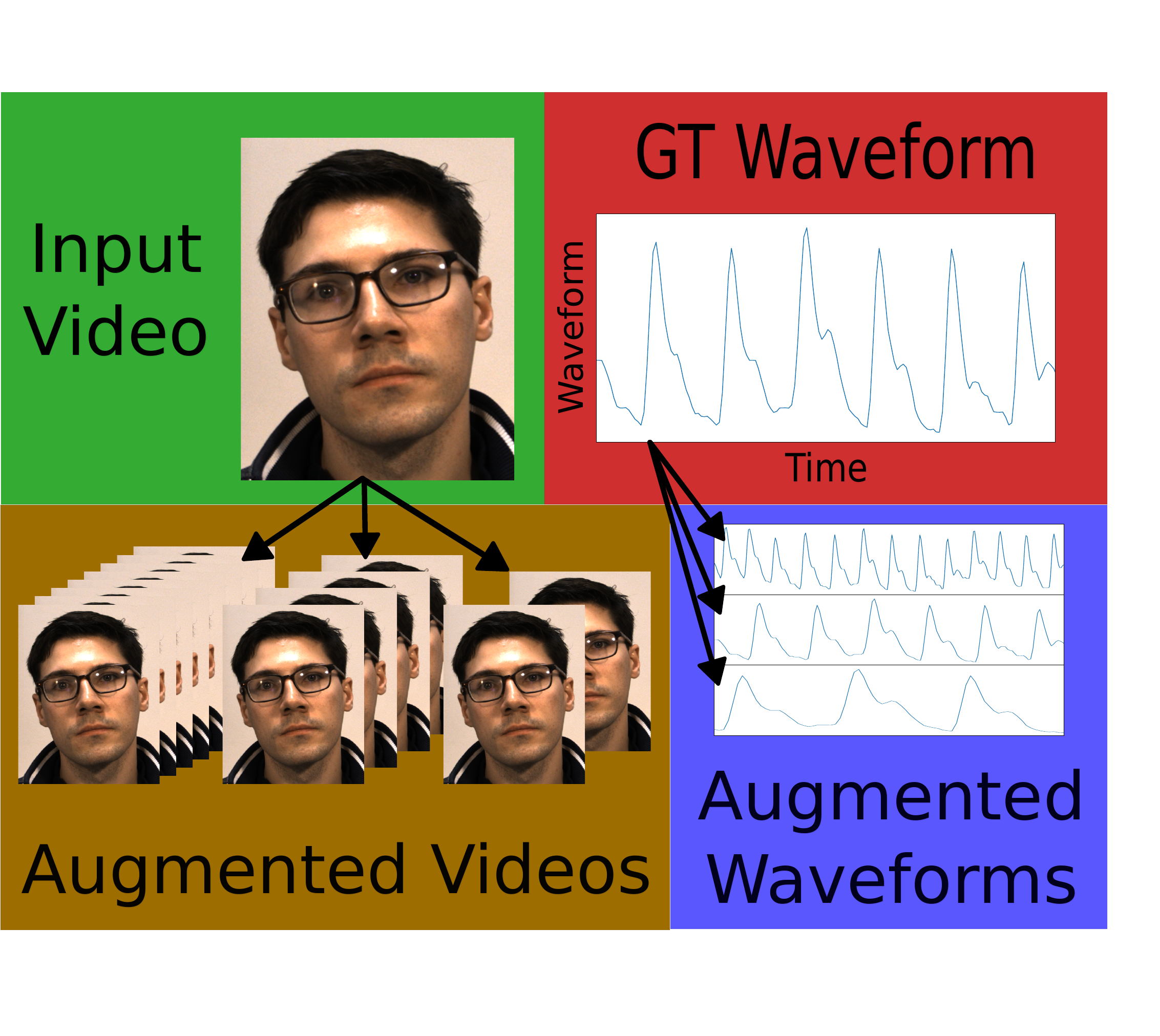}
    \caption{Overview of proposed temporal augmentations for rPPG. We interpolate both the training video and the waveform in order to train over a uniform distribution of heart rates.}
    \label{fig:overview}
\end{figure}

Training of rPPG models incorporates various types of data augmentations in the spatial domain. In this paper, we contribute a simple but very effective idea of augmenting the data in the temporal domain --- injecting synthetic data representing a wide spectrum of heart rates, thus allowing models to better respond to unknown heart rates. We evaluate this approach in a challenging cross-dataset setup comprising significant differences between heart rates in the training and test subsets. An overview of our augmentations targeting the temporal domain is shown in Figure \ref{fig:overview}.


\section{Related Work}

There has been broad interest in rPPG, with applications including detection of heart arrhythmias such as atrial fibrillation~\cite{sun2022contactless}, deepfake detection~\cite{qi2020deeprhythm}, and affective computing~\cite{sabour2021ubfc}.

Verkruysse \etal is credited with developing the first rPPG system, which relied on manually defined regions of interest, extraction of the green color channel, and applying a bandpass filter~\cite{verkruysse2008remote}. Poh \etal applied blind source separation and Independent Component Analysis (ICA) to boost performance~\cite{poh2010non}. Early techniques were not robust to motion, so de Haan and Jeanne developed CHROM, a motion-robust chrominance based rPPG system~\cite{DeHaan2013}. Wang \etal developed an rPPG system which projects color data to a ``plane orthogonal to the skin'' (POS), which further relaxes assumptions made with CHROM regarding subject skin tone~\cite{Wang2017}. Hsu \etal developed a support vector regression technique to predict the heart rate directly from rPPG features derived from Poh's ICA based method and CHROM~\cite{hsu2014learning}.

The emergence of practical deep learning methods has enabled new methods for rPPG estimation. Chen and McDuff developed DeepPhys, a CNN model based on VGG which effectively predicts pulse waveform derivatives based on adjacent video frames~\cite{chen2018deepphys}. Yu \etal developed a 3DCNN based approach for predicting the pulse waveform from video data~\cite{Yu2019}.

Cross-dataset generalization is a common concern with deep learning techniques, specifically in that deep learning rPPG techniques tend to perform suboptimally when working outside of the heart rate range of the training set~\cite{song2021pulsegan}. Tsou \etal developed Siamese-rPPG, a Siamese network utilizing 3D convolutions over two separate regions of interest, showing that this technique generalizes for cross-dataset analysis~\cite{tsou2020siamese}. Song \etal developed PulseGAN, a GAN based technique for generating more realistic PPG signals from the rPPG signals produced by CHROM, finding that this technique boosts performance even across datasets~\cite{song2021pulsegan}. Lu \etal expanded on this technique with Dual-GAN, which jointly predicts a realistic PPG signal and its noise distribution, and show improved cross-dataset performance as a result~\cite{lu2021dual}. In this paper, we develop \textit{speed} and \textit{modulation augmentations} for 3DCNN based models, showing that this consideration mitigates much of the cross dataset performance loss experienced by this family of models.
\section{Methods}

\begin{figure*}
    \centering
    \includegraphics[width=.8\linewidth]{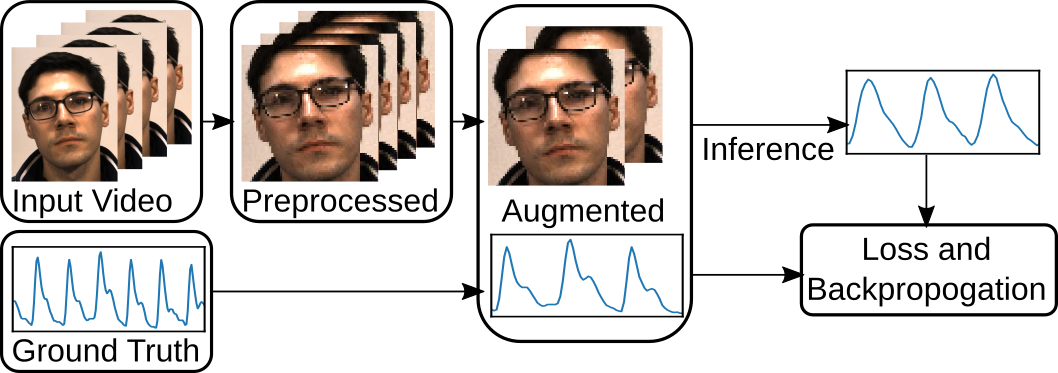}
    \caption{Overview of the temporal augmentation method. We apply the augmentations to the preprocessed data, then infer over the augmented images, and utilize the augmented waveform for calculating the negative Pearson loss.}
    \label{fig:method}
\end{figure*}

For rPPG analysis, we utilize the RPNet architecture~\cite{speth2021unifying}, which is a 3DCNN-based approach~\cite{Yu2019}. In particular, the network architecture is composed of 3D convolutions with max and global pooling layers for dimension reduction. The network consumes $64\times64$ pixel video over a 136-frame window, outputting an rPPG signal of 136 samples. In this section, we outline our video preprocessing and postprocessing steps, the training augmentations we employ, and other training parameters.

\subsection{Preprocessing and Postprocessing} \label{sec:prepost}

Our preprocessing pipeline consists of the following steps:

\begin{enumerate}
    \item We obtain facial landmarks at each frame in the dataset using the MediaPipe Face Mesh~\cite{lugaresi2019mediapipe} tool.
    \item We crop around the face at the extreme points of the landmarks, padded by 30\% on the top and 5\% on the sides and bottom, and the shortest dimension is extended to make the crop square.
    \item We scale the cropped portion to $64\times64$ pixels using cubic interpolation.
\end{enumerate}

When we perform a cross-dataset analysis, we reduce the frame rate of all videos to the lowest common denominator, \ie 30 FPS. This only affects the DDPM~\cite{speth2021deception} dataset, which is recorded at 90 FPS. The conversion takes place before the cropping step by taking the average pixel value over sets of three frames. We use this ``averaging'' technique rather than skipping frames as in~\cite{speth2021unifying} in order to better emulate a slower camera shutter speed.

RPNet outputs rPPG waves in 136-frame chunks with a stride of 68 frames. These parameters were selected so that the model would be small enough to fit on our GPUs. To reduce edge effects, we apply a Hann window to the overlapping segments and add them together, thus producing a single waveform.

As our evaluation protocol requires inferred heart rates, we take the Short-Time Fourier Transform (STFT) of the output waveform with a window size of 10 seconds and a stride of 1 frame, thus enabling the use of our system in application scenarios tolerant of a 10-second latency. We pad the waveform with zeros such that the bin width in the frequency domain is 0.001 Hz (0.06 beats per minute (BPM)) to reduce quantization effects. We select the highest peak in the range of $.6\overline{6}$ and 3 Hz (\ie 40 and 180 BPM) as the inferred heart rate.

\subsection{Augmentations}

We augment the temporal aspect of the training data, affecting alternatively the heart rate or \textit{speed}, and the change in heart rate or \textit{modulation}. An overview of our temporal augmentation framework showing how it fits into the training protocol is shown in Figure \ref{fig:method}.

To apply the speed augmentation, we first randomly select a target heart rate between 40 and 180 BPM (\ie the desired range of heart rates for which the model will be sensitive). We set this to be the same range as the peak selection used in the postprocessing step so that the model will be trained to predict the same heart rates that the rest of the system is designed to handle.

Second, we leverage the ground truth heart rate (obtained using the same STFT technique outlined in Section \ref{sec:prepost}), averaged over the 136 frame clip, as the source heart rate. We then calculate the length of data centered on the source clip to be $\lfloor 136 \times HR_{target} / HR_{source} \rfloor$.

Third, we interpolate the data in the source interval such that it becomes 136 frames long. This process is applied to both the video clip and the ground truth waveform.

To apply the modulation augmentation, we randomly select a modulation factor $f$ based on the ground truth heart rate such that when the clip speeds up or slows down by a factor of $f$, the change in heart rate is no more than 7 BPM per second. This parameter was selected based on the maximum observed change in heart rate in the DDPM dataset. We furthermore constrain the modulation such that the clip is modulated linearly by the selected factor over its duration, \ie for normalized heart rates $s$ and $e$ at the start and end of the clip respectively, the normalized heart rate at each frame $x$ in the $n$ frame clip (set to 136 as in Section \ref{sec:prepost}) is:

\begin{equation}
    nHR(x) = s + \frac{x(e-s)}{n}
\end{equation}

where $s = \frac{2}{1+f}$ and $e = sf$. We then integrate $nHR$ to generate a function yielding the positions $P(x)$ along the original clip at which to interpolate:

\begin{equation}
    P(x) = xs + \frac{x^2(e-s)}{2n} + c
\end{equation}

where $c=0$ due to indexing starting at 0. Finally, we linearly interpolate the $n$ frames from the original clip at every position $P(x)$ for all $x$ in the range $[0..n]$, thus yielding the modulated clip.

We additionally employ the horizontal flip, illumination, and Gaussian noise spatial augmentations from \cite{speth2021unifying}.

\subsection{Metrics}

We use the metrics proposed in~\cite{speth2021unifying} for our evaluation.
These metrics utilize either the pulse waveform (provided as ground truth or inferred by RPNet) or the heart rate (as derived in Section \ref{sec:prepost}). If the lengths of the ground truth and predicted waves differ (as is the case if the ground truth wave is not a multiple of 68 frames, \ie the stride used for RPNet), then we remove data points from the end of the ground truth wave such that they have the same length.

Each evaluation metric is calculated over each video in the dataset independently, the results of which are averaged. The following sections describe the evaluation metrics used in our experiments.

\subsubsection{Mean Error (ME)}

The ME captures the bias of the method in BPM, and is defined as follows:

\begin{equation}
    ME = \frac{1}{N}\sum\limits_{i=1}^{N} (HR'_i - HR_i)
\end{equation}

Where $HR$ and $HR'$ are the ground truth and predicted heart rates, respectively, where each contained index is the heart rate obtained from the STFT window as specified in Section \ref{sec:prepost}, and $N$ is the number of STFT windows present.

Many rPPG methods omit an analysis based on ME since it is often close to zero due to positive and negative errors canceling each other out. However, we find that it is valuable for gauging the bias of a model in a cross-dataset analysis by explaining how the model is failing, \ie whether the predictions are simply noisy or if they are shifted relative to the ground truth.

\subsubsection{Mean Absolute Error (MAE)}

The MAE captures an aspect of the precision of the method in BPM, and is defined as follows:

\begin{equation}
    MAE = \frac{1}{N}\sum\limits_{i=1}^{N} |HR'_i - HR_i|
\end{equation}

\subsubsection{Root Mean Squared Error (RMSE)}

The RMSE is similar to MAE, but penalizes outlier heart rates more strongly:

\begin{equation}
    RMSE = \sqrt{\frac{1}{N}\sum\limits_{i=1}^{N} (HR'_i - HR_i)^2}
\end{equation}

\subsubsection{Waveform Correlation ($r_{wave}$)}

The waveform correlation, $r_{wave}$, is the Pearson's $r$ correlation coefficient between the ground truth and predicted waves. When performing an inter-dataset analysis, we further maximize the $r_{wave}$ value by varying the correlation lag between ground truth and predicted waves by up to 1 second (30 data points) in order to compensate for differing synchronization techniques between datasets.
\section{Datasets} \label{sec:datasets}

For cross dataset analysis we utilized three rPPG datasets, chosen to contain a wide range of heart rates: PURE~\cite{stricker2014non}, UBFC-rPPG~\cite{bobbia2019unsupervised}, and DDPM~\cite{speth2021deception}. Key statistics for these three datasets are summarized in Table \ref{tab:datasets}.

\begin{table}
    \caption{Average duration, heart rate (HR) in BPM calculated using the STFT settings in Section \ref{sec:prepost}, and average within-session standard deviation in HR within a 60 second window and a stride of 1 frame, for PURE~\cite{stricker2014non}, UBFC-rPPG~\cite{bobbia2019unsupervised}, and DDPM~\cite{speth2021deception}. The 95\% confidence intervals are calculated across sessions in the dataset.}
    \centering\footnotesize
    \begin{tabular}{cccc}
        \toprule
        Dataset & Duration (s) & HR Avg & HR SD \\
        \midrule
        PURE & 68.307 $\pm$ 1.502 & 69.200 $\pm$ 6.026 & 1.638 $\pm$ 0.2682 \\
        UBFC & 64.964 $\pm$ 1.516 & 100.801 $\pm$ 5.056 & 3.016 $\pm$ 0.525 \\
        DDPM & 656.464 $\pm$ 22.310 & 96.982 $\pm$ 4.186 & 4.000 $\pm$ 0.286 \\
        \bottomrule
    \end{tabular}
    \label{tab:datasets}
\end{table}

\subsection{PURE}

The PURE dataset is useful for cross-dataset analysis for two key reasons. First, it has the lowest average heart rate of the three datasets, being about 30 BPM lower than the other two. Second, it has the lowest within-subject heart rate standard deviation.

\subsection{UBFC-rPPG}

The UBFC-rPPG dataset (in this paper shortened to UBFC) features subjects playing a time-sensitive mathematical game which caused a heightened physiological response. UBFC has the highest average heart rate of the three datasets
and more heart rate variability than PURE, but less variability than DDPM.

\subsection{DDPM}

The DDPM dataset is the largest of the compared datasets, with recorded sessions lasting nearly 11 minutes on average. It also features the most heart rate variability of the three, with a heart rate standard deviation of about 4 BPM. This is due to stress-inducing aspects (mock interrogation with forced deceptive answers) in the collection protocol of DDPM.
Due to noise in the ground truth oximeter waveforms, we mask out all 10 second segments in DDPM where the heart rate changes by more than 7 BPM per second.
\section{Training}

\begin{figure*}
    \centering
    \begin{subfigure}{.33\textwidth}
        \centering
        \includegraphics[width=\textwidth]{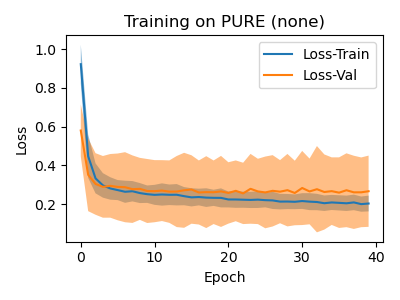}
        \caption{}
        \label{fig:pure-none}
    \end{subfigure}
    \begin{subfigure}{.33\textwidth}
        \centering
        \includegraphics[width=\textwidth]{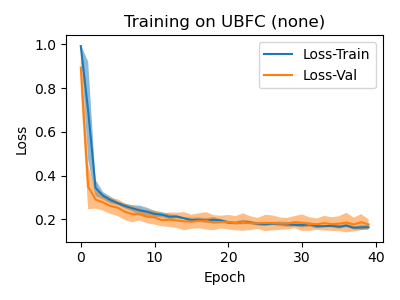}
        \caption{}
        \label{fig:ubfc-none}
    \end{subfigure}
    \begin{subfigure}{.33\textwidth}
        \centering
        \includegraphics[width=\textwidth]{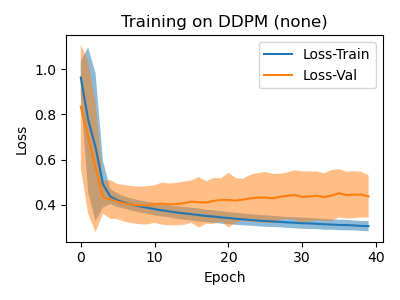}
        \caption{}
        \label{fig:ddpm-none}
    \end{subfigure}
    \begin{subfigure}{.33\textwidth}
        \centering
        \includegraphics[width=\textwidth]{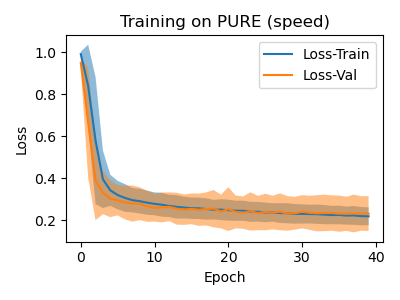}
        \caption{}
        \label{fig:pure-linear}
    \end{subfigure}
    \begin{subfigure}{.33\textwidth}
        \centering
        \includegraphics[width=\textwidth]{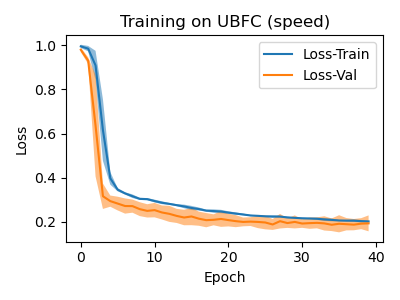}
        \caption{}
        \label{fig:ubfc-linear}
    \end{subfigure}
    \begin{subfigure}{.33\textwidth}
        \centering
        \includegraphics[width=\textwidth]{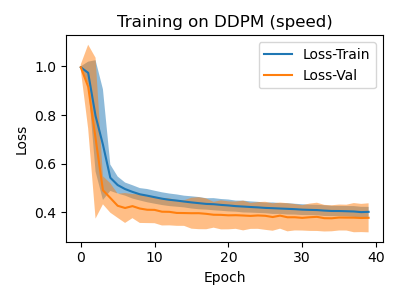}
        \caption{}
        \label{fig:ddpm-linear}
    \end{subfigure}
    \begin{subfigure}{.33\textwidth}
        \centering
        \includegraphics[width=\textwidth]{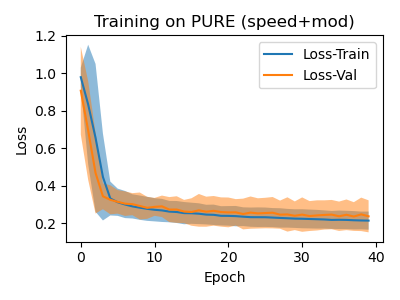}
        \caption{}
        \label{fig:pure-fourier}
    \end{subfigure}
    \begin{subfigure}{.33\textwidth}
        \centering
        \includegraphics[width=\textwidth]{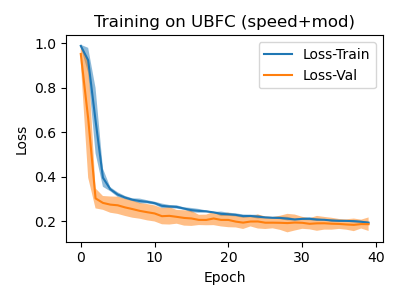}
        \caption{}
        \label{fig:ubfc-fourier}
    \end{subfigure}
    \begin{subfigure}{.33\textwidth}
        \centering
        \includegraphics[width=\textwidth]{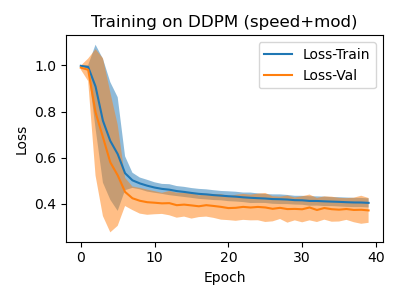}
        \caption{}
        \label{fig:ddpm-fourier}
    \end{subfigure}
    \caption{Training RPNet on PURE, UBFC, and DDPM, utilizing no temporal augmentations, speed, and speed plus modulation augmentations.}
    \label{fig:training}
\end{figure*}

For each of the three datasets, we randomly partition the videos into five subject-disjoint sets, three of which are merged to generate splits for training, validation, and testing at 3/1/1 ratios. We then rotate the splits to generate five folds for cross-validation. We train for 40 epochs using the negative Pearson loss function~\cite{Yu2019} and the Adam optimizer configured with a 0.0001 learning rate. Models are selected based on minimum validation loss.

Figure \ref{fig:training} shows training and validation losses when training RPNet on the three datasets outlined in Section~\ref{sec:datasets} and applying three augmentation settings: none, speed, and speed+mod. We observe that utilizing any sort of temporal augmentation causes the validation loss to converge with tighter confidence intervals. This is especially evident when training on the PURE dataset where the median validation loss confidence interval without temporal augmentations (Figure \ref{fig:pure-none}) drops from $\pm0.174$ to $\pm0.081$ and $\pm0.078$ with speed and speed+mod augmentations, respectively (Figures~\ref{fig:pure-linear} and~\ref{fig:pure-fourier}). Furthermore, while it is apparent from Figure~\ref{fig:ddpm-none} that training over DDPM without temporal augmentations can lead to overfitting, both temporal augmentation settings appear to avoid this problem (Figures~\ref{fig:ddpm-linear} and~\ref{fig:ddpm-fourier}).

Across all combinations of augmentations and datasets, the validation loss converges to a lower value when temporal augmentations are used than when they are not. We believe that this is because the models are forced to generalize when the range and variability of heart rates they are exposed to is increased, limiting the effectiveness of simply memorizing a signal which looks like a heart rate and replaying it at a frequency common to the dataset.

\section{Experimental Results} \label{sec:results}

We trained and tested RPNet on each of the three datasets discussed in Section~\ref{sec:datasets}, both in a within-dataset analysis (3 training-testing configurations with PURE-PURE, UBFC-UBFC, and DDPM-DDPM), and with a cross-dataset analysis (6 training-testing configurations with PURE-UBFC, PURE-DDPM, UBFC-PURE, UBFC-DDPM, DDPM-PURE, and DDPM-UBFC). Furthermore, we investigated 3 temporal augmentation settings, namely no temporal augmentation (none), speed augmentation (speed), and speed plus modulation augmentation (speed+mod). The results for the within-dataset analysis are shown in Table~\ref{tab:withinds} and for the cross-dataset analysis are shown in Table~\ref{tab:xdsresults}.

\begin{table*}
    \caption{Results for the 9 within-dataset combinations of dataset and the temporal augmentations used. Heart rate metrics (ME, MAE, and RMSE) have units of BPM, and $r_{wave}$ is Pearson's r correlation over pulse waveforms.}
    \centering\footnotesize
    \begin{tabular}{cccccc}
        \toprule
        Dataset & Augmentations & ME & MAE & RMSE & $r_{wave}$ \\
        \midrule
        PURE & none & -0.516 $\pm$ 1.814 & 1.176 $\pm$ 1.891 & 1.872 $\pm$ 3.067 & 0.694 $\pm$ 0.253 \\
        PURE & speed & -0.012 $\pm$ 0.461 & 0.694 $\pm$ 0.566 & 1.222 $\pm$ 1.456 & \textbf{0.753 $\pm$ 0.087} \\
        PURE & speed+mod & \textbf{0.006 $\pm$ 0.389} & \textbf{0.639 $\pm$ 0.482} & \textbf{1.130 $\pm$ 1.347} & 0.752 $\pm$ 0.089 \\
        \midrule
        UBFC & none & 0.922 $\pm$ 2.215 & 1.432 $\pm$ 2.201 & 2.238 $\pm$ 2.630 & \textbf{0.803 $\pm$ 0.024} \\
        UBFC & speed & \textbf{0.016 $\pm$ 0.384} & 0.616 $\pm$ 0.201 & 1.346 $\pm$ 0.746 & 0.793 $\pm$ 0.020 \\
        UBFC & speed+mod & 0.091 $\pm$ 0.139 & \textbf{0.502 $\pm$ 0.121} & \textbf{0.993 $\pm$ 0.335} & 0.798 $\pm$ 0.024 \\
        \midrule
        DDPM & none & -1.443 $\pm$ 5.725 & 4.167 $\pm$ 4.680 & 6.907 $\pm$ 6.504 & 0.569 $\pm$ 0.070 \\
        DDPM & speed & \textbf{-0.773 $\pm$ 2.036} & 3.230 $\pm$ 2.267 & 5.897 $\pm$ 4.671 & 0.584 $\pm$ 0.052 \\
        DDPM & speed+mod & -1.048 $\pm$ 1.434 & \textbf{2.981 $\pm$ 1.738} & \textbf{5.485 $\pm$ 3.412} & \textbf{0.587 $\pm$ 0.057} \\
        \bottomrule
    \end{tabular}
    \label{tab:withinds}
\end{table*}

\begin{table*}
    \caption{Results for the 18 cross-dataset combinations of train dataset, test dataset, and temporal augmentations used. Heart rate metrics (ME, MAE, and RMSE) have units of BPM, while $r_{wave}$ is Pearson's r correlation over pulse waveforms.}
    \centering\footnotesize
    \begin{tabular}{ccccccc}
        \toprule
        Train & Test & Augmentations & ME & MAE & RMSE & $r_{wave}$ \\
        \midrule
        PURE & UBFC & none & -13.082 $\pm$ 12.972 & 13.690 $\pm$ 12.847 & 19.320 $\pm$ 13.359 & 0.532 $\pm$ 0.136 \\
        PURE & UBFC & speed & -3.340 $\pm$ 2.998 & 4.703 $\pm$ 3.083 & 9.219 $\pm$ 4.645 & 0.590 $\pm$ 0.102 \\
        PURE & UBFC & speed+mod & \textbf{-1.491 $\pm$ 0.583} & \textbf{2.251 $\pm$ 0.671} & \textbf{5.191 $\pm$ 1.559} & \textbf{0.636 $\pm$ 0.053} \\
        \midrule
        PURE & DDPM & none & -27.633 $\pm$ 8.058 & 32.360 $\pm$ 3.934 & 38.397 $\pm$ 3.052 & 0.182 $\pm$ 0.015 \\
        PURE & DDPM & speed & -10.926 $\pm$ 11.184 & \textbf{24.343 $\pm$ 4.140} & \textbf{33.410 $\pm$ 3.694} & \textbf{0.221 $\pm$ 0.032} \\
        PURE & DDPM & speed+mod & \textbf{6.436 $\pm$ 4.870} & 33.620 $\pm$ 2.018 & 42.494 $\pm$ 2.829 & 0.150 $\pm$ 0.015 \\
        \midrule
        \midrule
        UBFC & PURE & none & 9.657 $\pm$ 3.971 & 11.532 $\pm$ 2.710 & 14.791 $\pm$ 2.751 & 0.619 $\pm$ 0.021 \\
        UBFC & PURE & speed & \textbf{0.864 $\pm$ 1.074} & \textbf{2.196 $\pm$ 0.921} & \textbf{3.758 $\pm$ 1.289} & \textbf{0.671 $\pm$ 0.043} \\
        UBFC & PURE & speed+mod & 0.938 $\pm$ 0.720 & 2.535 $\pm$ 0.920 & 4.246 $\pm$ 1.275 & 0.625 $\pm$ 0.025 \\
        \midrule
        UBFC & DDPM & none & -5.569 $\pm$ 4.479 & \textbf{14.947 $\pm$ 2.231} & \textbf{20.738 $\pm$ 2.366} & \textbf{0.264 $\pm$ 0.028} \\
        UBFC & DDPM & speed & \textbf{-4.240 $\pm$ 6.961} & 18.574 $\pm$ 2.707 & 28.082 $\pm$ 3.056 & 0.251 $\pm$ 0.020 \\
        UBFC & DDPM & speed+mod & 11.258 $\pm$ 4.904 & 32.914 $\pm$ 0.769 & 41.698 $\pm$ 0.834 & 0.174 $\pm$ 0.010 \\
        \midrule
        \midrule
        DDPM & PURE & none & 26.092 $\pm$ 14.065 & 26.660 $\pm$ 13.435 & 30.915 $\pm$ 13.164 & 0.437 $\pm$ 0.099 \\
        DDPM & PURE & speed & \textbf{1.256 $\pm$ 1.563} & \textbf{2.208 $\pm$ 1.824} & \textbf{3.905 $\pm$ 2.996} & \textbf{0.686 $\pm$ 0.061} \\
        DDPM & PURE & speed+mod & 1.338 $\pm$ 1.477 & 2.509 $\pm$ 1.776 & 4.441 $\pm$ 2.991 & 0.673 $\pm$ 0.058 \\
        \midrule
        DDPM & UBFC & none & \textbf{-0.358 $\pm$ 0.863} & 1.963 $\pm$ 1.135 & 3.745 $\pm$ 1.931 & 0.699 $\pm$ 0.050 \\
        DDPM & UBFC & speed & -0.431 $\pm$ 0.177 & 1.311 $\pm$ 0.282 & 3.140 $\pm$ 0.654 & 0.711 $\pm$ 0.028 \\
        DDPM & UBFC & speed+mod & -0.563 $\pm$ 0.383 & \textbf{1.160 $\pm$ 0.393} & \textbf{2.906 $\pm$ 1.112} & \textbf{0.734 $\pm$ 0.029} \\
        \bottomrule
    \end{tabular}
    \label{tab:xdsresults}
\end{table*}

While the temporal augmentations were intended to improve cross-dataset performance, we did observe a slight performance boost in the within-dataset case. As shown in Table~\ref{tab:withinds}, all metrics except $r_{wave}$ on UBFC exhibited better performance when temporal augmentations were employed. However, in these cases the performance boost is slight, often falling within the 95\% confidence intervals of the results without augmentation.

Our primary interest is in the cross-dataset case shown in Table~\ref{tab:xdsresults}. We found that training on a dataset with higher heart rate variability and testing on a dataset with lower heart rate variability tends to produce better results than the reverse. This is especially evident in cross dataset cases involving DDPM, which has the highest heart rate variability as measured by heart rate standard deviation in Table~\ref{tab:datasets}.

We were particularly interested in the cross-dataset performance between the relatively low heart rate dataset PURE and the higher heart rate datasets DDPM and UBFC. As shown in the ME column of Table~\ref{tab:xdsresults}, we observe that when training and testing between datasets of different heart rates without temporal augmentations, the bias as reflected by ME is strong, with UBFC-PURE yielding the ME closest to zero at over 9 BPM. Furthermore, these models are biased in the direction of the training dataset's mean heart rate, \ie training on PURE which has relatively low heart rates results in a negative ME on UBFC and DDPM, while training on UBFC or DDPM results in a positive ME when testing on PURE. However, applying the speed augmentation causes ME to be much closer to zero than when no such augmentation is used. This is because the speed augmentation is intended to mitigate the heart rate bias inherent in the training dataset, thus causing it to generalize to any heart rates seen in the augmented training regime rather than simply those present in the dataset. With the mitigation of heart rate bias as reflected by improved ME scores, we observe an improvement in MAE and RMSE in most cases. We furthermore observe a boost in $r_{wave}$, indicating that the models more faithfully reproduce the waveforms with low noise.

The modulation augmentation is intended to boost performance when training on a dataset with low heart rate variability such as PURE and testing on a dataset with high variability such as UBFC and DDPM. We observe that modulation indeed boosts performance for PURE-UBFC, though even with modulation PURE-DDPM fails to generalize. With the possible exception of DDPM-UBFC, we do not observe the modulation augmentation positively impacting cases when the training dataset already contains high heart rate variability, as is the case with UBFC and DDPM.

\begin{table}
    \caption{Zero-effort errors obtained by predicting the average heart rate of the dataset for all subjects. In all cases ME is 0.}
    \centering\footnotesize
    \begin{tabular}{ccc}
        \toprule
        Dataset & MAE & RMSE \\
        \midrule
        PURE & 15.847 & 23.054 \\
        UBFC & 14.085 & 17.256 \\
        DDPM & 17.804 & 22.113 \\
        \bottomrule
    \end{tabular}
    \label{tab:averagehrErrors}
\end{table}

We observe poor results in both cross dataset experiments where DDPM is the test dataset. Of those, we still observe the same trend in PURE-DDPM as we observe in other cases, \ie that models trained with speed augmentations outperform those without, albeit in this case the performance is still quite poor. In UBFC-DDPM we see that models trained without speed augmentations achieve better results than with speed augmentations, which is a break from the trend observed in all other cases. Furthermore, whereas in other cases high MAE and RMSE errors are largely explained by bias as reflected in ME, this case has a relatively low ME relative to MAE and RMSE. We believe that in this case since the average heart rate between UBFC and DDPM is relatively close (differing by less than 4 BPM), overfitting to this band of heart rates is actually beneficial for the cross dataset analysis. Furthermore, we investigated the ``zero-effort'' 
error rates achieved by a model which simply predicts the average heart rate for the dataset (97 BPM as in Table~\ref{tab:datasets}), finding comparable error rates to UBFC-DDPM (MAE and RMSE are 17.804 and 22.113 respectively). These zero-effort
results for the three datasets are reported in Table~\ref{tab:averagehrErrors}.

\begin{figure}
    \centering
    \includegraphics[width=\linewidth]{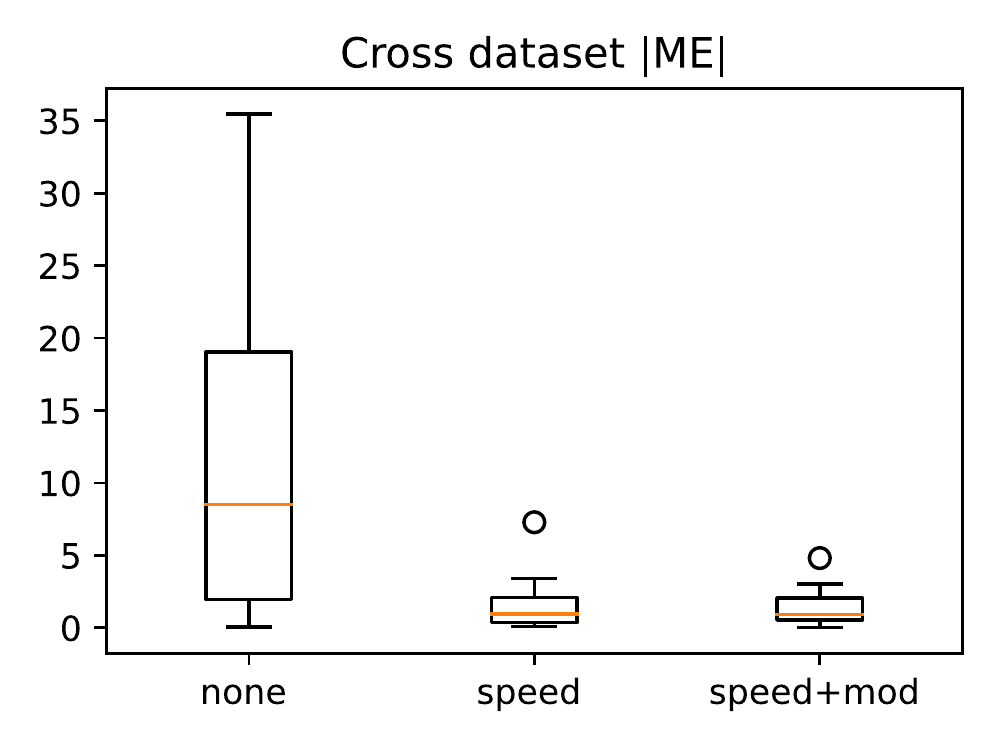}
    \caption{Speed augmentations reduce learned bias as reflected by a reduced $|$ME$|$ in cross dataset analysis between datasets with differing heart rate bands.}
    \label{fig:xdssummaries-box-me}
\end{figure}

\begin{figure}
    \centering
    \includegraphics[width=\linewidth]{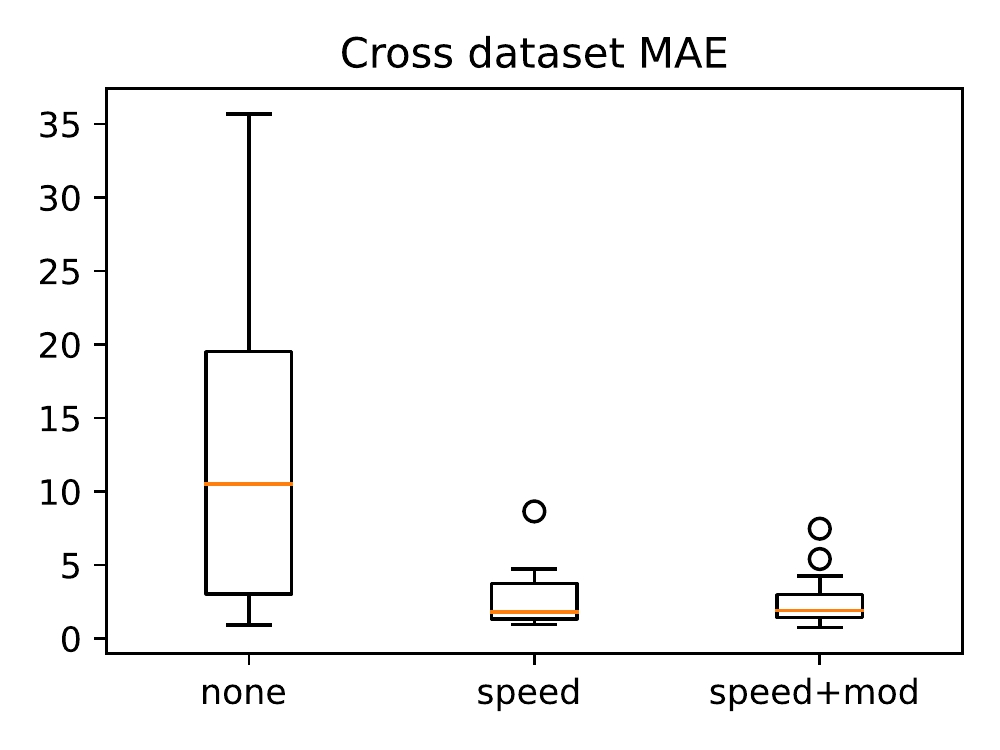}
    \caption{Speed augmentations can improve the accuracy of the model, reflected by an improved MAE.}
    \label{fig:xdssummaries-box-mae}
\end{figure}

We summarise the cross dataset results in Table~\ref{tab:xdssummaries}. In this case we calculate the 95\% confidence interval across 4 cross dataset combinations (omitting the cases when testing on DDPM as no models generalized) and 5 training folds. We find that combining both speed and modulation losses yields optimal performance on all metrics. The box plots in Figures~\ref{fig:xdssummaries-box-me} and~\ref{fig:xdssummaries-box-mae} further demonstrate the reason why the temporal augmentations outperform the case without augmentations. In particular, the bias of the model to predict heart rates similar to its training dataset has been significantly reduced, as is most clearly seen in the reduced absolute ME shown in Figure~\ref{fig:xdssummaries-box-me}. We further observe an improved MAE shown in Figure~\ref{fig:xdssummaries-box-mae}.

\begin{table*}
    \caption{Summaries of cross dataset performance under speed augmentation settings, omitting PURE-DDPM and UBFC-DDPM where no models succeed in generalizing. We take the absolute value of ME metrics before averaging.}
    \centering\footnotesize
    \begin{tabular}{ccccc}
        \toprule
        Augmentations & $|$ME$|$ & MAE & RMSE & $r_{wave}$ \\
        \midrule
        none & 12.349 $\pm$ 5.546 & 13.460 $\pm$ 5.335 & 17.192 $\pm$ 5.720 & 0.572 $\pm$ 0.056 \\
        speed & 1.502 $\pm$ 0.803 & 2.604 $\pm$ 0.884 & 5.005 $\pm$ 1.536 & 0.664 $\pm$ 0.031 \\
        speed+mod & \textbf{1.373 $\pm$ 0.570} & \textbf{2.501 $\pm$ 0.784} & \textbf{4.830 $\pm$ 1.174} & \textbf{0.677 $\pm$ 0.025} \\
        \bottomrule
    \end{tabular}
    \label{tab:xdssummaries}
\end{table*}


\begin{table}
    \caption{We compare RPNet to other methods: CHROM~\cite{DeHaan2013}, POS~\cite{Wang2017}, Siamese-rPPG~\cite{tsou2020siamese}, PulseGAN~\cite{song2021pulsegan}, and Dual-GAN~\cite{lu2021dual}. Because postprocessing steps differ between published methods, we perform our analysis of RPNet with several postprocessing settings.}
    \centering\footnotesize
    \begin{tabular}{ccccc}
        \toprule
        Train & Test & Method & MAE & RMSE \\
        \midrule
        NA & PURE & CHROM & 2.237 & 4.697 \\
        NA & PURE & POS & 2.609 & 5.532 \\
        UBFC & PURE & Siamese-rPPG & 0.63 & 2.51 \\
        UBFC & PURE & RPNet-$w_{10}$ & 2.251 $\pm$ 0.671 & 5.191 $\pm$ 1.559 \\
        UBFC & PURE & RPNet-$w_{30}$ & 0.741 $\pm$ 0.121 & 1.592 $\pm$ 0.207 \\
        UBFC & PURE & RPNet-$w_{full}$ & 0.958 $\pm$ 0.073 & 2.349 $\pm$ 0.125 \\
        \midrule
        NA & UBFC & CHROM & 3.114 & 6.136 \\
        NA & UBFC & POS & 3.363 & 7.366 \\
        PURE & UBFC & Siamese-rPPG & 1.29 & 8.73 \\
        PURE & UBFC & PulseGAN & 2.09 & 4.42  \\
        PURE & UBFC & Dual-GAN & 0.74 & 1.02 \\
        PURE & UBFC & RPNet-$w_{10}$ & 2.535 $\pm$ 0.920 & 4.246 $\pm$ 1.275 \\
        PURE & UBFC & RPNet-$w_{30}$ & 1.925 $\pm$ 1.163 & 2.797 $\pm$ 1.326 \\
        PURE & UBFC & RPNet-$w_{full}$ & 1.480 $\pm$ 0.707 & 4.939 $\pm$ 4.002 \\
        \bottomrule
    \end{tabular}
    \label{tab:comp_other_methods}
\end{table}

We compare our method with other methods in the rPPG literature. 
Several factors contribute uncertainty to this analysis:

\begin{itemize}
    \item The Siamese-rPPG method does not include settings for calculating the FFT spectrogram for heart rate derivation, which as argued in~\cite{mironenko2020remote} can introduce uncertainty into the comparison with this method.
    \item Both GAN based methods use interbeat intervals to derive the heartrate, which differs from our method which relies on an STFT specrogram.
    \item PulseGAN is trained on both PURE and BSIPL-RPPG (an in-house database), whereas RPNet was trained without BSIPL-RPPG.
    \item The GAN techniques solve a somewhat different problem in that they use CHROM signals as an input in order to generate a waveform with more realistic PPG features, whereas the others infer the pulse waveform from video data.
\end{itemize}




To compensate for these differences, we evaluate the RPNet models trained using speed and modulation augmentations under three different postprocessing configurations: 1) $w_{10}$ uses the 10-second STFT window as described in~\ref{sec:prepost}; 2) $w_{30}$ uses a 30 second STFT window, but otherwise leaves the evaluation the same; 3) $w_{full}$ calculates the FFT over the full waveform, and results across all subjects are concatenated before calculating the RMSE metric. The results are shown in Table~\ref{tab:comp_other_methods}.

While it is unclear (given the variety of postprocessing steps) how our method ranks compared to other rPPG techniques, for the more lenient configurations the results show a MAE within the $\pm2$ BPM or $\pm2\%$ published accuracy bounds of CMS50E series oximeters (used in the collection of the PURE, UBFC-rPPG, and DDPM datasets). Furthermore, we believe that our recommended augmentations are generally applicable to deep learning based rPPG as a whole, as this augmentation strategy may be implemented as a training framework for any model architecture that trains based on video inputs to produce waveform outputs.

\section{Conclusions}

In this paper, we show the importance of temporal speed-based augmentations for the cross-dataset generalization of deep learning rPPG methods. We develop a system for training deep learning rPPG models using two variants of this augmentation method, \ie speed augmentation affecting the heart rate, and modulation affecting the change in heart rate. We argue that these augmentations may be applied to any deep learning rPPG system which produces a pulse waveform from video inputs.

While this paper probed an interesting failure case of deep learning in rPPG, much room for improvement remains. We were unable to achieve satisfactory performance training on the relatively simple PURE or UBFC datasets and testing on the more complex DDPM dataset, likely due to extreme head pose changes and dynamic facial expressions spurred by the interrogation collection setting of DDPM. It is conceivable that a set of augmentations targeting spatial distortion can permit generalization in these dimensions, which future work should investigate.

We found cross dataset performance to be comparable to other published work. However, due to differences in postprocessing steps which have little to no bearing on the performance of the algorithm itself, we were unable to perform a full and comprehensive comparison. We believe that the effect of postprocessing on rPPG should be studied and recommendations made for the community to standardize on common techniques.

{\small
\bibliographystyle{ieee_fullname}
\bibliography{main}

\begin{thebibliography}{10}\itemsep=-1pt

\bibitem{bobbia2019unsupervised}
Serge Bobbia, Richard Macwan, Yannick Benezeth, Alamin Mansouri, and Julien
  Dubois.
\newblock Unsupervised skin tissue segmentation for remote
  photoplethysmography.
\newblock {\em Pattern Recognition Letters}, 124:82--90, 2019.

\bibitem{chen2018deepphys}
Weixuan Chen and Daniel McDuff.
\newblock Deepphys: Video-based physiological measurement using convolutional
  attention networks.
\newblock In {\em Proceedings of the european conference on computer vision
  (ECCV)}, pages 349--365, 2018.

\bibitem{DeHaan2013}
G. {de Haan} and V. {Jeanne}.
\newblock Robust pulse rate from chrominance-based rppg.
\newblock {\em IEEE Trans. on Biom. Eng.}, 60(10):2878--2886, 2013.

\bibitem{hsu2014learning}
YungChien Hsu, Yen-Liang Lin, and Winston Hsu.
\newblock Learning-based heart rate detection from remote photoplethysmography
  features.
\newblock In {\em 2014 IEEE International Conference on Acoustics, Speech and
  Signal Processing (ICASSP)}, pages 4433--4437. IEEE, 2014.

\bibitem{lin2019face}
Bofan Lin, Xiaobai Li, Zitong Yu, and Guoying Zhao.
\newblock Face liveness detection by rppg features and contextual patch-based
  cnn.
\newblock In {\em Proceedings of the 2019 3rd international conference on
  biometric engineering and applications}, pages 61--68, 2019.

\bibitem{liu2020multi}
Xin Liu, Josh Fromm, Shwetak Patel, and Daniel McDuff.
\newblock Multi-task temporal shift attention networks for on-device
  contactless vitals measurement.
\newblock {\em Advances in Neural Information Processing Systems},
  33:19400--19411, 2020.

\bibitem{lu2021dual}
Hao Lu, Hu Han, and S~Kevin Zhou.
\newblock Dual-gan: Joint bvp and noise modeling for remote physiological
  measurement.
\newblock In {\em Proceedings of the IEEE/CVF Conference on Computer Vision and
  Pattern Recognition}, pages 12404--12413, 2021.

\bibitem{lugaresi2019mediapipe}
Camillo Lugaresi, Jiuqiang Tang, Hadon Nash, Chris McClanahan, Esha Uboweja,
  Michael Hays, Fan Zhang, Chuo-Ling Chang, Ming~Guang Yong, Juhyun Lee, et~al.
\newblock Mediapipe: A framework for building perception pipelines.
\newblock {\em arXiv preprint arXiv:1906.08172}, 2019.

\bibitem{mironenko2020remote}
Yuriy Mironenko, Konstantin Kalinin, Mikhail Kopeliovich, and Mikhail
  Petrushan.
\newblock Remote photoplethysmography: Rarely considered factors.
\newblock In {\em Proceedings of the IEEE/CVF Conference on Computer Vision and
  Pattern Recognition Workshops}, pages 296--297, 2020.

\bibitem{poh2010non}
Ming-Zher Poh, Daniel~J McDuff, and Rosalind~W Picard.
\newblock Non-contact, automated cardiac pulse measurements using video imaging
  and blind source separation.
\newblock {\em Optics express}, 18(10):10762--10774, 2010.

\bibitem{qi2020deeprhythm}
Hua Qi, Qing Guo, Felix Juefei-Xu, Xiaofei Xie, Lei Ma, Wei Feng, Yang Liu, and
  Jianjun Zhao.
\newblock Deeprhythm: Exposing deepfakes with attentional visual heartbeat
  rhythms.
\newblock In {\em Proceedings of the 28th ACM international conference on
  multimedia}, pages 4318--4327, 2020.

\bibitem{sabour2021ubfc}
Rita~Meziati Sabour, Yannick Benezeth, Pierre De~Oliveira, Julien Chappe, and
  Fan Yang.
\newblock Ubfc-phys: A multimodal database for psychophysiological studies of
  social stress.
\newblock {\em IEEE Transactions on Affective Computing}, 2021.

\bibitem{song2021pulsegan}
Rencheng Song, Huan Chen, Juan Cheng, Chang Li, Yu Liu, and Xun Chen.
\newblock Pulsegan: Learning to generate realistic pulse waveforms in remote
  photoplethysmography.
\newblock {\em IEEE Journal of Biomedical and Health Informatics},
  25(5):1373--1384, 2021.

\bibitem{speth2021deception}
Jeremy Speth, Nathan Vance, Adam Czajka, Kevin~W Bowyer, Diane Wright, and
  Patrick Flynn.
\newblock Deception detection and remote physiological monitoring: A dataset
  and baseline experimental results.
\newblock In {\em 2021 IEEE International Joint Conference on Biometrics
  (IJCB)}, pages 1--8. IEEE, 2021.

\bibitem{speth2021unifying}
Jeremy Speth, Nathan Vance, Patrick Flynn, Kevin Bowyer, and Adam Czajka.
\newblock Unifying frame rate and temporal dilations for improved remote pulse
  detection.
\newblock {\em Computer Vision and Image Understanding}, 210:103246, 2021.

\bibitem{stricker2014non}
Ronny Stricker, Steffen M{\"u}ller, and Horst-Michael Gross.
\newblock Non-contact video-based pulse rate measurement on a mobile service
  robot.
\newblock In {\em The 23rd IEEE International Symposium on Robot and Human
  Interactive Communication}, pages 1056--1062. IEEE, 2014.

\bibitem{sun2022contactless}
Yu Sun, Yin-Yin Yang, Bing-Jhang Wu, Po-Wei Huang, Shao-En Cheng, Bing-Fei Wu,
  and Chun-Chang Chen.
\newblock Contactless facial video recording with deep learning models for the
  detection of atrial fibrillation.
\newblock {\em Scientific reports}, 12(1):1--10, 2022.

\bibitem{tsou2020siamese}
Yun-Yun Tsou, Yi-An Lee, Chiou-Ting Hsu, and Shang-Hung Chang.
\newblock Siamese-rppg network: Remote photoplethysmography signal estimation
  from face videos.
\newblock In {\em Proceedings of the 35th annual ACM symposium on applied
  computing}, pages 2066--2073, 2020.

\bibitem{verkruysse2008remote}
Wim Verkruysse, Lars~O Svaasand, and J~Stuart Nelson.
\newblock Remote plethysmographic imaging using ambient light.
\newblock {\em Optics express}, 16(26):21434--21445, 2008.

\bibitem{Wang2017}
W. {Wang}, A.~C. {den Brinker}, S. {Stuijk}, and G. {de Haan}.
\newblock Algorithmic principles of remote ppg.
\newblock {\em IEEE Trans. on Biom. Eng.}, 64(7):1479--1491, 2017.

\bibitem{Yu2019}
Zitong Yu, Xiaobai Li, and Guoying Zhao.
\newblock Remote photoplethysmograph signal measurement from facial videos
  using spatio-temporal networks.
\newblock In {\em British Machine Vision Conf.}, 2019.

\end{thebibliography}
}

\end{document}